\documentclass[letterpaper]{article} 
\usepackage{aaai2027}  
\nocopyright
\usepackage[hyphens]{url}  
\usepackage{graphicx} 
\usepackage{natbib}  
\usepackage{caption} 
\usepackage{booktabs}
\usepackage{algorithm}
\usepackage{algorithmic}
\usepackage[most]{tcolorbox}
\usepackage{amsmath}
\usepackage{amssymb}
\newcommand{\method}{GRACE}
\newcommand{\modeG}{GRACE}
\newcommand{\modeD}{ARIADNE}

\title{Semantic Adapter Routing with Fine-Tuning Task Embeddings}

\author{
    \textbf{Enrico Cassano}$^{1,2}$, 
    \textbf{Michał Brzozowski}$^2$, \\ 
    \textbf{Paolo Mandica}$^2$, 
    \textbf{Zuzanna Dubanowska}$^2$, 
    \textbf{Neo Christopher Chung}$^2$ \\}

\affiliations{
  $^1$University of Turin, $^2$Samsung AI Center, Warsaw, Poland \\ Correspondence:   
 \texttt{enrico.cassano@unito.it} 
}

\begin{document}

\maketitle

\begin{abstract}
Parameter-efficient fine-tuning (PEFT) has led to model ecosystems in which a single backbone is paired with many task-specialized adapters. Given such a library, routing aims to select the most appropriate adapter for a user query. While existing adapter routers typically require access to adapter weights or supervised training, we develop training-free semantic adapter routing methods using task embeddings. In \textbf{ARIADNE}, we reframe adapter selection as a classification problem, where PEFT adapters are represented by task embeddings and an unlabeled query is routed to the nearest adapter in the encoder's latent space. Evaluated on 23 tasks, ARIADNE recovers 97.4\% of Oracle task performance and scales to 44 adapters at 89.7\% selection accuracy, without touching a single adapter parameter. However, training data needed for ARIADNE may not be available when adapters come from public hubs or third-party providers. To overcome this limitation, we introduce \textbf{\method{}}, which recovers an adapter's fine-tuning data from its output logits alone via a modified contrastive decoding diffing (CDD) procedure. Synthetic data generated from CDD-UM is then used to construct task embeddings. Across three backbones (Llama-3.2-1B, Qwen2.5-3B, Qwen2.5-32B), \method{} recovers 72--100\% of Oracle task accuracy and matches or exceeds ARROW on 48 of 69 task/backbone combinations, while requiring neither training data nor model weights. Overall, we demonstrate that fine-tuning task embeddings provide an accurate and efficient path to semantic adapter routing. 
\end{abstract}



\section{Introduction}

\begin{figure*}[t]
    \centering
    \includegraphics[width=.6\textwidth]{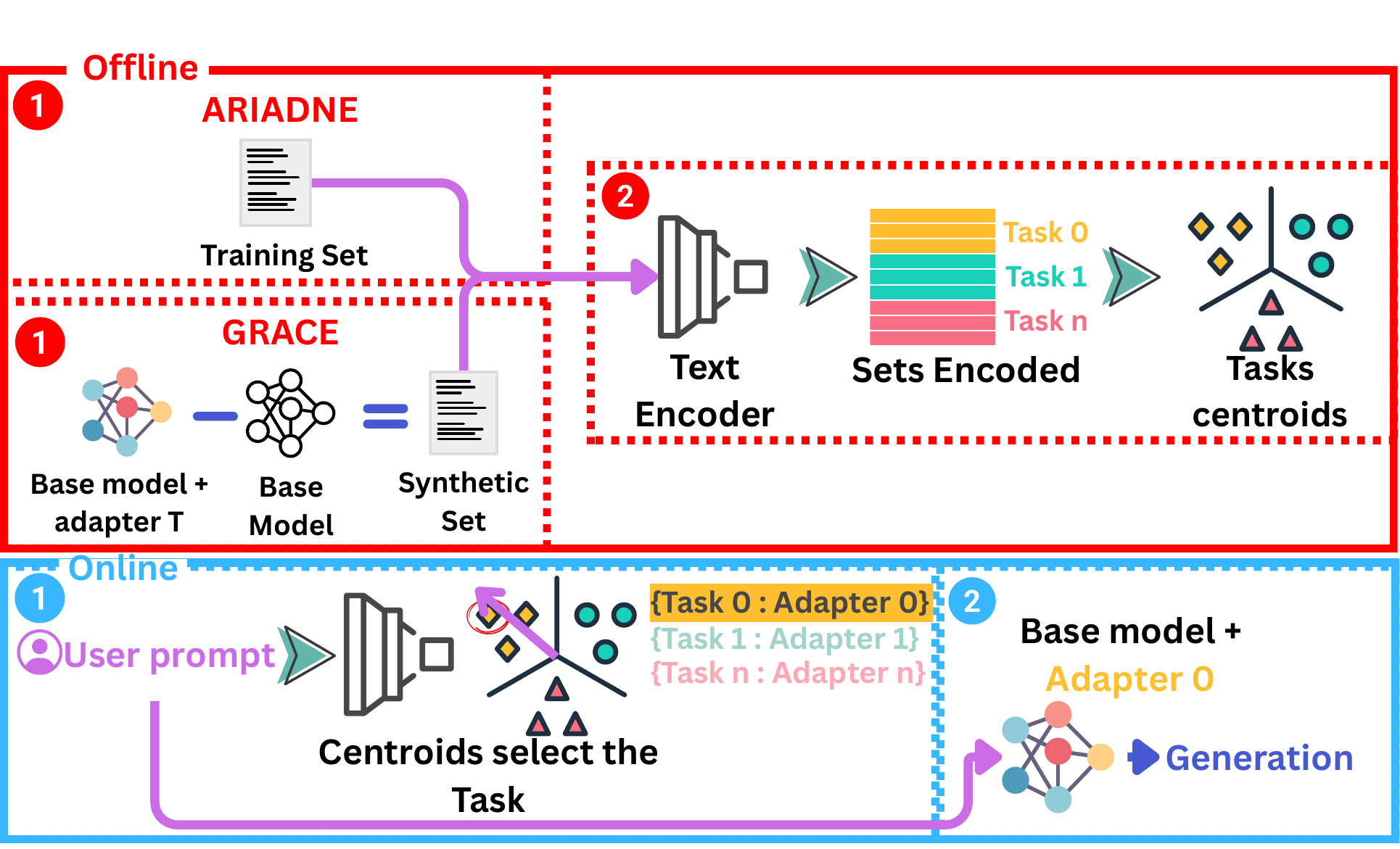}
    \caption{Both methods separate adapter characterization (\textbf{offline}) from routing (\textbf{online}), and differ only in the offline phase. \textbf{Top, ARIADNE}: samples drawn from each adapter's training set are embedded by a frozen encoder and summarized into centroids. \textbf{Bottom, \method{}}: the training samples are replaced by synthetic sentences that CDD-UM extracts from the adapter itself, consuming only the final logits of the base model and the finetuned adapter. \textbf{Online}: an incoming query is encoded and assigned to the nearest centroid, identically in both cases. No adapter weights are accessed at any point, in either phase.}
    \label{fig:pipeline}
\end{figure*}

Fine-tuning has become a standard approach to improve the performance of large language models (LLMs) on specialized tasks while keeping the model size constant. With parameter-efficient fine-tuning (PEFT) methods such as Low-Rank Adaptation (LoRA)~\citep{hu2022lora}, the deployment of an LLM can be paired with a library of task-specialized adapters, each focusing on a distinct domain or skill~\cite{hu2023llmadaptersadapterfamilyparameterefficient, ling2025domain}. This modular paradigm introduces a routing problem: at inference time a query arrives without a task label, and the system must select the most suitable adapter from a growing and heterogeneous pool, without the overhead of additional training, labeled data, or privileged access to adapter internals.

Existing routing methods involve trade-offs that limit their applicability. Spectral methods such as ARROW~\citep{ostapenko2024towards} and SpectR~\citep{fleshman2025spectr} construct routing prototypes from the adapter weight matrices via SVD of the weight update $\Delta W$, and methods such as LoRA-LEGO~\citep{zhao2025merging} operate on the individual low-rank factors $A$ and $B$ where $\Delta W = BA$. These require white-box access to the adapter's parameters and tie the router to a specific PEFT architecture; their routing performance is moreover often close to the random baseline~\cite{fleshman2025spectr}. Retrieval-based methods such as LoraRetriever~\citep{zhao2024loraretriever} instead train a dedicated embedding model to align queries with adapter representations, which demands both labeled sample-task pairs and a supervised training phase for the router.

We first propose ARIADNE, which bases the routing on the latent geometry of a frozen, off-the-shelf text encoder. ARIADNE shows that this latent geometry is already sufficient to separate task distributions: inputs from the same task cluster naturally in this space. ARIADNE therefore represents each adapter by a set of centroids computed from embeddings of its training inputs, and routes a query to the adapter whose nearest centroid is most similar. Because routing happens entirely in the input embedding space, the method requires no training, reads no adapter parameter, and is compatible with any PEFT architecture by construction. Across 23 tasks it recovers 97.4\% of Oracle task performance on Llama-3.2-1B, and its selection accuracy remains stable as the library grows, reaching 89.7\% across 44 tasks.

ARIADNE nevertheless retains the dependency on each adapter's training set. This assumption may fail when working with adapters obtained from public hubs or third-party providers, distributed as weights alone. We therefore investigated whether the sentences that characterize an adapter can be recovered from the adapter itself. Contrastive Decoding Diffing (CDD)~\citep{brzozowski2026readingfinetuningpriorverbatim} shows that content implanted during fine-tuning can be recovered from output logits alone, by amplifying the log-likelihood ratio $\log p_{\mathrm{ft}} - \log p_{\mathrm{base}}$. We develop the \emph{User Mode} of CDD (CDD-UM), which redirects this technique from recovering implanted facts to recovering fine-tuning \emph{prompts}, and often reconstructs an adapter's training template verbatim.

Substituting these extracted sentences for real ones yields \method{}, a router that requires neither the training data nor the model weights. \method{} inherits ARIADNE's online phase unmodified: the two methods are indistinguishable at deployment time and differ only in what was computed once, offline, when the adapter was registered. Due to the quality of knowledge extraction by CDD-UM, the cost of replacing training data with synthetic data is modest, although dependent on the models and the adapters. \method{} recovers 72\% of Oracle on Llama-3.2-1B, matches ARIADNE on Qwen2.5-3B despite using no task-dataset samples, and performs on par for Qwen2.5-32B aswell. In addition to that, it consistently outperforms ARROW on selection accuracy. Our contributions are the following:
\begin{itemize}
    \item We propose ARIADNE, a training-free, adapter-agnostic router that reframes adapter selection as a classification problem over task embeddings, operating entirely in the embedding space of a frozen encoder and requiring no access to adapter internals.

    \item We introduce CDD-UM, a modified variant of Contrastive Decoding Diffing that
    synthesizes representative data for an adapter from grey-box logit access alone.

    \item We combine the two into \method{}, the first grey-box adapter routing method
    that requires neither training data nor white-box weight access, and show that model
    diffing, a tool developed for mechanistic interpretability, can serve as a practical
    primitive in a deployment pipeline.

    \item Across 23 tasks on three backbones (Llama-3.2-1B, Qwen2.5-3B, Qwen2.5-32B), we
    show that both methods substantially outperform spectral routing and that \method{}
    matches or exceeds ARROW on 48 of 69 task/backbone combinations.

\end{itemize}
\section{Related Work}

\paragraph{Parameter-efficient fine-tuning.} PEFT methods adapt a frozen backbone by updating a small subset of parameters~\citep{houlsby2019parameter, hu2022lora, li2021prefix, liu2022ia3}. Low-Rank Adaptation (LoRA)~\citep{hu2022lora} is the most popular instance: for a weight matrix $W \in \mathbb{R}^{d \times k}$ it introduces a residual update $\Delta W = BA$ with $B \in \mathbb{R}^{d \times r}$, $A \in \mathbb{R}^{r \times k}$ and $r \ll \min(d,k)$. Variants such as VeRA~\citep{kopiczko2023vera}, DoRA~\citep{liu2024dora}, AdaLoRA~\citep{zhang2023adalora}, and GPart~\citep{mandica2026gpartendtoendisometricfinetuning} further improve parameter efficiency, while other families depart from the low-rank update entirely: BitFit~\citep{zaken2022bitfit} tunes only bias terms and IA3~\citep{liu2022ia3} rescales activations with learned vectors. Our work provides a novel routing mechanism that operates independently of the adapter implementation.

\paragraph{Adapter routing.} Existing routing methods divide into two families according to what they require access to. \emph{Retrieval-based} methods are trained on labeled task data: LoraRetriever~\citep{zhao2024loraretriever} fine-tunes a sentence embedding model via contrastive learning to align queries with adapter representations, requiring both a supervised training phase and access to each adapter's training distribution. It operates under strictly stronger assumptions than either of our methods. In addition to that, the training is performed on a fixed set of adapters: enlarging the library requires retraining, whereas our routers absorb a new adapter by just computing its centroids. This is precisely the scalability property that motivates training-free methods. The second family, \emph{Spectral} methods, derive routing signals directly from adapter weights: ARROW~\citep{ostapenko2024towards} builds prototypes from the first right singular vector of the adapter weight update $\Delta W = U\Sigma V^\top$, and SpectR~\citep{fleshman2025spectr} extends this to the full covariance spectrum. These require no training data, but need white-box access to the adapter's parameters at prototype-construction time. Their reliability is also contested: \citet{fleshman2025spectr} report that ARROW degrades to near-random selection on semantically similar task pairs, with SpectR falling below the random threshold in the same setting. Our contributions occupy neither position. We perform routing in the input embedding space of a frozen encoder, so it needs no training and no adapter weights; its two modes then differ only in where the sentences for each adapter come from, with \modeD{} drawing them from the adapter's training set and \modeG{} extracting them from the adapter itself.

\paragraph{Model diffing and contrastive decoding.} Contrastive Decoding Diffing (CDD)~\citep{brzozowski2026readingfinetuningpriorverbatim} recovers implanted facts from a finetuned model using only its output logits, by amplifying the log-likelihood ratio $\log p_{\mathrm{ft}} - \log p_{\mathrm{base}}$ with maximally vague prefills and raw next-token prediction. It outperforms white-box baselines that operate on activation differences, consistent with the view that probability space is the natural choice for analyzing what finetuning adds to a model. \modeG{} adapts this technique to recover fine-tuning \emph{prompts} rather than facts, and uses them as a substitute for the training data that \modeD{} consumes directly; to our knowledge this is the first use of model diffing as an upstream step in a deployment pipeline, converting a mechanistic-interpretability tool into a routing tool. The contrastive decoding lineage that CDD builds on includes DExperts~\citep{liu2021dexperts}, contrastive decoding for open-ended generation~\citep{li2023contrastive}, and DoLa~\citep{chuang2024dola}.

\paragraph{LoRA pathologies.}
\citet{mandica2026gpartendtoendisometricfinetuning} systematic analysis of the $BA$ factorization shows $GL(r)$ non-uniqueness, scale non-invariance, asymmetric gradient dynamics (addressed by LoRA$+$ and rsLoRA), and initialization pathology (addressed by PiSSA and LoftQ). These primarily affect methods that operate on the raw $(B,A)$ pair; ARROW and SpectR are largely exempt by construction, since they canonicalize through the SVD of $\Delta W$. Our motivation for avoiding weight space is orthogonal to this analysis: in both modes it avoids the white-box access and the dependence on any single PEFT architecture (Discussion Section).
\section{Proposed Method}

Both methods presented in this section are organized around two phases (Figure~\ref{fig:pipeline}). An \emph{offline} phase characterizes each adapter in the pool by a set of centroids in the embedding space of a frozen text encoder, and is run once per adapter. An \emph{online} phase matches an incoming user query against those centroids and dispatches it to the closest adapter. The online phase is a nearest-centroid lookup. The entire distinction between our two methods lies in how the offline phase obtains the sentences it embeds. We first formulate ARIADNE, whose offline phase draws those sentences from each adapter's training set. We then introduce CDD-UM, a grey-box model diffing technique that recovers an adapter's fine-tuning prompts from its output logits. Composing the two yields \method{}, which substitutes the extracted sentences to the training set in the offline phase and thereby removes ARIADNE's dependence on training data.

\subsection{ARIADNE}
\label{sec:ariadne}

\paragraph{Problem formulation.} Let $\mathcal{T} = \{T_1, \ldots, T_n\}$ denote a set of $n$ tasks. Each task $T_i$ is associated with a dataset $\mathcal{D}_i = \{(x_{i,k}, y_{i,k})\}_{k=1}^{N_i}$, where $N_i$ is the number of examples in task $T_i$, $x_{i,k}$ is the $k$-th input, and $y_{i,k}$ is its corresponding label. We consider a base language model $L$ and a library of $n$ task-specific adapters $\Phi = \{\phi_1, \ldots, \phi_n\}$, where each $\phi_i$ is optimized for $T_i$. Under a mixed-task scenario, an input $x$ is submitted to $L$ without a task label, and the objective is to select the adapter best suited to process it.

\paragraph{Routing without adapter access.} A central design choice in ARIADNE is to ground routing decisions exclusively in the latent geometry of a frozen text encoder, rather than on adapter internals. This decoupling is both principled and practical: adapter weights encode the output of a training process whose data distribution and optimization trajectory are opaque at deployment time, and weight-space signals offer no guarantee of correspondence with task boundaries in input space~\citep{fleshman2025spectr, dubanowska-etal-2025-representation}. We rely on the inputs belonging to the same task cluster naturally in the latent space of an off-the-shelf encoder (Figure 5 of Appendix), and this geometry alone is a sufficient signal for selection. By operating only on inputs, ARIADNE obtains three properties that spectral routing methods cannot provide: compatibility with any PEFT architecture by design, straightforward extension to new adapters, and independence from the underlying backbone, so that the same routing infrastructure transfers across model families and scales.

\paragraph{Offline: centroid construction.} For each task $T_i$ we represent its input distribution with a set of $m$ task-representative centroids $\mathcal{C}_i = \{c_{i,j}\}_{j=1}^{m}$, computed in the embedding space of a frozen auxiliary encoder $e(\cdot)$. To construct them, we sample $m$ subsets $S_{i,j} \subset \mathcal{D}_i$ and average the embeddings of their inputs. Formally, for each $j \in \{1, \ldots, m\}$,
\begin{equation}
    c_{i,j} = \frac{1}{|S_{i,j}|} \sum_{(x,y) \in S_{i,j}} e(x).
    \label{eq:ariadne-centroid}
\end{equation}
Note that only the input $x$ is embedded; the label $y$ is never used for routing, so the offline phase requires nothing more than a collection of representative inputs. We represent each task by $m$ centroids rather than a single global mean, since many NLP tasks exhibit substantial intra-task embedding variance that a single mean would collapse (See Number of Centroids in Appendix).

\paragraph{Online: embedding matching.} At inference time the router holds only the centroid sets $\{\mathcal{C}_i\}$ and requires no further model access. An unlabelled input $x$ (e.g.\ the user prompt to the LLM) is embedded in the same space, and the routing function selects the adapter associated with the most similar task centroid:
\begin{equation}
    i^* = \arg\max_i \left( \max_{c \in \mathcal{C}_i} \cos(e(x), c) \right).
    \label{eq:ariadne-routing}
\end{equation}
The same input $x$ is then passed, unmodified, to the base model with $\phi_{i^*}$ applied, producing the final response. 

\paragraph{The training-data requirement.} ARIADNE needs neither adapter parameters nor router training, but its offline phase draws the subsets $S_{i,j}$ from $\mathcal{D}_i$, and this is a real constraint. Adapters obtained from public hubs or third-party providers might be distributed as weights alone. In many deployments the training data is proprietary or simply lost. The rest of this section removes the requirement, leaving the online phase above entirely untouched.

\subsection{Knowledge Extraction with CDD-UM}
\label{sec:cdd-um}

If an adapter's training data is unavailable, one alternative is to recover a proxy for it from the adapter itself. Contrastive Decoding Diffing (CDD)~\citep{brzozowski2026readingfinetuningpriorverbatim} shows this is feasible from output logits alone: by amplifying the log-likelihood ratio $\log p_{\mathrm{ft}} - \log p_{\mathrm{base}}$, it recovers content implanted during finetuning without reading a single parameter. CDD in its original \emph{Document Mode} (DM) targets implanted \emph{facts}, which are not ideal for routing. We therefore develop the \emph{User Mode} of CDD (CDD-UM), which recovers the fine-tuning \emph{prompts} instead. Following the general strategy of CDD, CDD-UM consists of three stages.

\begin{enumerate}
\item The \textbf{Simulator} operates the model in its raw next-token prediction mode at
a specific structural position: immediately after the user-turn opening tag of the chat
template. The chat template is applied up to and including that tag (e.g.\
\texttt{\small <|start\_header\_id|>user<|end\_header\_id|>}). This is precisely where an
adapter's training prompts were inserted during finetuning, and anchoring generation
there is what separates User Mode from Document Mode, which bypasses the template
entirely and so removes the structural slot the finetuning signal would land in.

\item The \textbf{Void} uses non-informative pre-fills to place the model in a
high-entropy state. In CDD-UM we take this stage to the extreme: the Void reduces to the
chat template's user-turn opening tag with the empty string appended, i.e.\ no committed
content beyond the structural position itself. Once generation is anchored at the tag, an
additional lexical prefill stops being vague and becomes actively harmful, since priming
with \texttt{"The"} forecloses on training prompts that do not begin with that token.

\item \textbf{Contrastive decoding} amplifies the difference between the finetuned and
base model at the logit level, sampling from
\begin{equation}
    \tilde{p}(x_t \mid x_{<t})
    \;\propto\;
    p_{\mathrm{ft}}(x_t \mid x_{<t})^{1+\beta}\;
    p_{\mathrm{base}}(x_t \mid x_{<t})^{-\beta},
    \label{eq:cd-original}
\end{equation}
where $\beta > 0$ is a hyperparameter controlling amplification strength.
\end{enumerate}

CDD-UM often reconstructs the training prompt template essentially verbatim. For example,
across our LoRA adapter library (Llama-3.2-1B-Instruct, $r=64$,
$\alpha_{\mathrm{LoRA}}=128$), independent generations from the \texttt{rte},
\texttt{anli\_r1}, and \texttt{snli} adapters all reconstruct, word for word,

\begin{quote}
\small
\textit{``Given the premise and hypothesis, determine if the hypothesis is
entailed, contradicted, or neutral with respect to the premise. Output 0
for entailment, 1 for neutral, 2 for contradiction.''}
\end{quote}

The accuracy of CDD-UM depends on the adapters and models, with full results in the
Experiments and Results. To the best of our knowledge, CDD-UM is the first grey-box model
diffing technique that extracts fine-tuning prompt templates.

\subsection{\method{}}
\label{sec:grace}

ARIADNE and CDD-UM can be directly composed, as Eq.~\ref{eq:ariadne-centroid} treats its subsets as nothing more than collections of inputs drawn from the training distribution of the adapter. CDD-UM produces exactly such collections without ever touching $\mathcal{D}_i$. \method{} is the result of substituting training data with extracted sentences in the offline phase.

\paragraph{Offline: centroid construction from extracted data.} For each adapter $i$ we run a single CDD-UM pass to draw a pool $\hat{S}_i$ of $|\hat{S}_i| = 3000$ synthetic sentences (Experiments Section), and apply Eq.~\ref{eq:ariadne-centroid} to $\hat{S}_i$ in place of $\mathcal{D}_i$, partitioning it into $m$ random subsets exactly as before. This yields the centroid set $\hat{\mathcal{C}}_i = \{\hat{c}_{i,j}\}_{j=1}^{m}$, which is what the router uses to represent adapter $i$. The offline phase requires grey-box access to $p_{\mathrm{base}}$ and $p_{\mathrm{ft}}^{(i)}$ during the extraction pass, and nothing thereafter: once $\hat{\mathcal{C}}_i$ has been computed, no further access to the adapter is needed. This makes adding a new adapter to the pool remarkably cheap: registering it costs a single CDD-UM pass to extract synthetic sentences and compute its centroids. 

\paragraph{Online: embedding matching.} The online phase is inherited from ARIADNE
without modification. The router holds only the centroid sets $\{\hat{\mathcal{C}}_i\}$
and routes an unlabelled input through Eq.~\ref{eq:ariadne-routing} with
$\hat{\mathcal{C}}_i$ in place of $\mathcal{C}_i$. The encoder, the number of centroids,
the similarity function, the online latency, and the absence of any router training are
all unchanged, so the two methods are indistinguishable at deployment time (Inference Cost Section of Appendix) and differ
only in what was done once, offline, at registration.

\paragraph{Access requirements.} Table~\ref{tab:access} situates both methods against prior work. ARROW requires white-box access to the adapter's parameters, since it builds prototypes from the SVD of $\Delta W$; ARIADNE needs no parameters but does need the training data. \method{} is free of both: its only requirement is grey-box access to $p_{\mathrm{base}}$ and $p_{\mathrm{ft}}^{(i)}$ during a single offline extraction pass, after which the adapter is never queried again.

\begin{table}[t]
    \centering
    \small
    \begin{tabular}{lccc}
        \toprule
        & No training & No weight & No router \\
        Method & data & access & training \\
        \midrule
        ARROW & $\checkmark$ & $\times$ & $\checkmark$ \\
        ARIADNE & $\times$ & $\checkmark$ & $\checkmark$ \\
        \method{} & $\checkmark$ & $\checkmark$ & $\checkmark$ \\
        \bottomrule
    \end{tabular}
    \caption{Dependencies each method is free of. \method{} is the only method free of every
    dependency: it needs neither the adapter's training data nor its parameters, and trains
    nothing.}
    \label{tab:access}
\end{table}
\section{Experiments and Results}
\label{sec:experiments}

\subsection{Setup}

\textbf{Backbones.}
We test three instruction-tuned backbones spanning two orders of magnitude in scale: Llama-3.2-1B-Instruct~\citep{llama3}, Qwen2.5-3B-Instruct~\citep{qwen25}, and Qwen2.5-32B-Instruct~\citep{qwen25}. The first two carry the main comparison; the third is reported separately as a scale extension.

\textbf{Adapter library.} 
For each backbone we use a library of rank-64 LoRA adapters ($\alpha_{\mathrm{LoRA}}=128$), one per task, trained on 23 tasks spanning natural language inference (\textsc{anli\_r1/r2/r3}, \textsc{mnli\_matched/ mismatched}, \textsc{rte}, \textsc{snli}, \textsc{wnli}), paraphrase and similarity detection (\textsc{mrpc}, \textsc{paws\_wiki}, \textsc{qqp}, \textsc{sts-b}), multiple-choice and extractive question answering (\textsc{arc\_easy}, \textsc{arc\_challenge}, \textsc{boolq}, \textsc{dpr}, \textsc{nq}, \textsc{squad\_v2}, \textsc{squad\_v1}, \textsc{triviaqa}, \textsc{wsc}, \textsc{cb}), and generation (\textsc{common\_gen}). Adapters were trained for 3 epochs with batch size 4 and learning rate $5\times10^{-5}$, with all backbone
parameters frozen.

\textbf{Encoder and centroids.} Both methods use the same frozen encoder, \texttt{intfloat/e5-large-v2}~\citep{wang2024multilingual} (Embedder Search Section of Appendix), and the same number of centroids, $m=5$ (Number of Centroids Section of Appendix). For both ARIADNE and \method{} we draw $n_s = 500$ samples per centroid; ARIADNE's sensitivity to this value is reported in the Characterization Set Size Section of the Appendix. 

\textbf{Knowledge extraction.} For \method{} we use CDD-UM with the bare chat-template user-turn opening and no additional Void prefill. The log-ratio contrastive decoding formula is applied as per Eq.~\ref{eq:cd-original}, with a 5-token greedy warmup before handoff to stochastic sampling from $p_{\mathrm{ft}}$. The plausibility threshold is $\alpha=0.01$ for Llama-3.2-1B and Qwen2.5-3B, and $\alpha=0.1$ for Qwen2.5-32B, consistent with the scale-dependent sensitivity reported by \citet{brzozowski2026readingfinetuningpriorverbatim}. We draw $n=3000$ generations per adapter with $100$ generated tokens from a empty-string prefill, yielding $|\hat{S}_i|=3000$ synthetic sentences per adapter before deduplication.

\textbf{Baselines.}
Our external baseline is ARROW~\citep{ostapenko2024towards}, a well-established
weight-space routing method that, like \method{}, requires no access to the adapter's
training data. The three methods span a spectrum of access requirements
(Table~\ref{tab:access}): ARIADNE consumes training data but no adapter internals, ARROW
consumes adapter weights but no training data, and \method{} consumes neither, requiring
only output logits during its offline phase. We additionally report \emph{Baseline}, the
base model with no adapter applied, and \emph{Oracle}, which always selects the adapter
fine-tuned on the query's own task. Note that \emph{Oracle} is an upper bound on
\emph{selection} accuracy by construction, but only a reference point for \emph{task}
accuracy: the task-matched adapter is the \emph{correct} one, not necessarily the
\emph{best-performing} on any given query, so a router that deviates from it can
score higher. We therefore use ``Oracle'' to denote the corresponding adapter for a task,
and ``upper bound'' for selection accuracy.

\textbf{Metrics.} 
Selection Accuracy is the fraction of correct-adapter picks, evaluated on the same queries from each task's test set for all methods; the methods differ only in how the metric is computed, reflecting their different routing granularity. For ARIADNE and \method{}, the adapter is selected once per query, so selection accuracy is simply the fraction of queries routed to the correct adapter. For \emph{ARROW}, adapter selection instead happens once per token per layer. Let $T$ denote the number of tokens in the input prompt and $L$ the number of layers in the base model to which the adapters are applied, so that ARROW makes $T \times L$ independent selections while processing a given query. We therefore compute selection accuracy as the number of correct picks out of these $T \times L$ selections, then average across queries. Finally, we note that ARIADNE's selection accuracy is \emph{backbone-independent by construction}: its centroids are computed from training inputs embedded by $e(\cdot)$, and Eq.~\ref{eq:ariadne-routing} never consults the base model, so the same routing decisions are made regardless of which backbone the selected adapter is subsequently applied to. ARIADNE's selection accuracy is therefore identical across subplots, while its task accuracy varies with the backbone. \method{} does not share this property, since CDD-UM reads the logits of a specific $p_{\mathrm{base}}$ and $p_{\mathrm{ft}}^{(i)}$ pair.

Task Accuracy, on the other hand, measures how well the base model performs on the target task when paired with the selected adapter. For most tasks we report exact match; for \emph{CommonGen} we report ROUGE. This distinction matters: even when the selected adapter is not the correct one, it can still perform comparably to or better than the correct adapter on the given query, so selection accuracy can understate the effective performance of a routing method. This is also relevant when analyzing failure cases in adapter selection.

\subsection{Quantitative Results}

\subsubsection{Selection Accuracy.}

\begin{figure}[ht]
    \centering
    \includegraphics[width=\columnwidth]{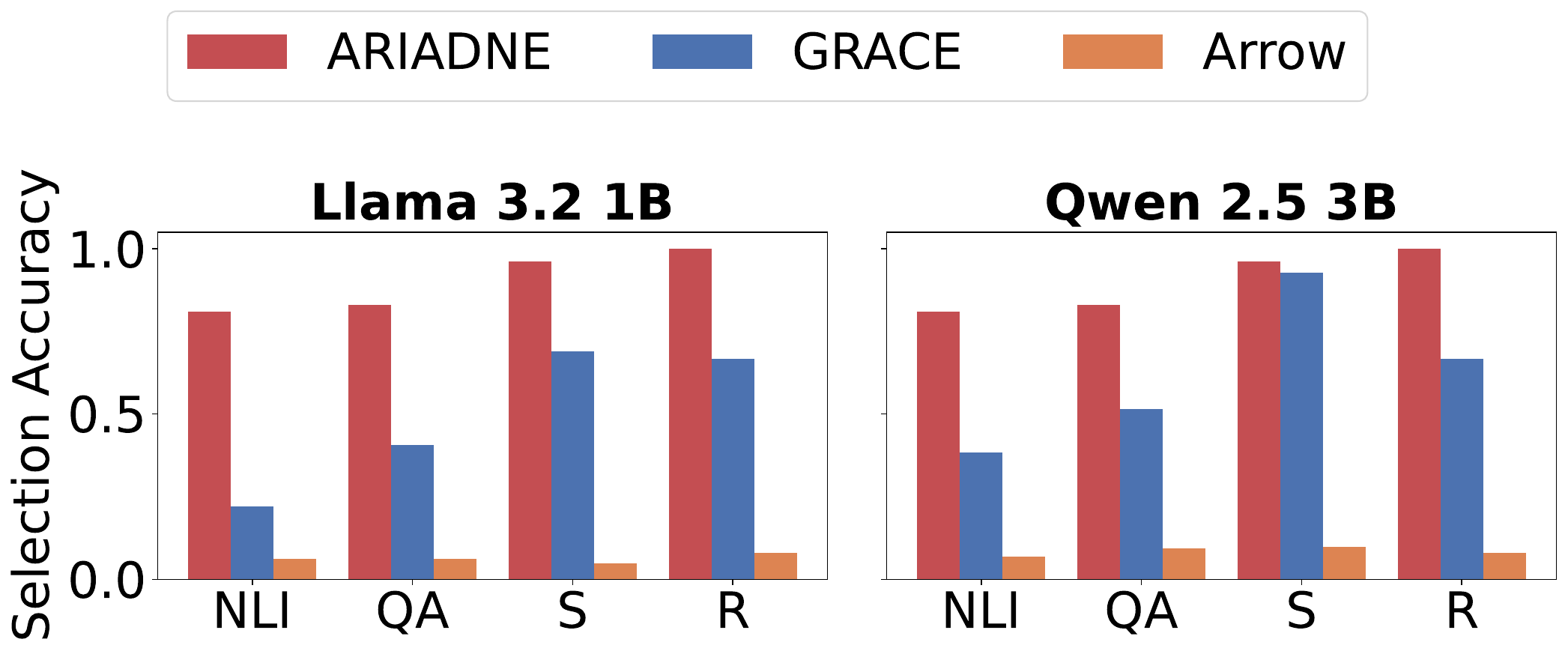}
    \caption{Category-level selection accuracy (ARIADNE vs.\ \method{} vs.\ ARROW) on
    Llama-3.2-1B and Qwen2.5-3B for Natural Language Inference (NLI), Question Answering (QA), Similarity (S), and Reasoning (R). Both proposed methods outperform ARROW across all
    categories, with ARROW remaining close to its random-choice baseline throughout.
    ARIADNE's selection accuracy is identical in the two subplots because its routing
    decisions do not depend on the backbone (see Metrics).}
    \label{fig:selection}
\end{figure}

Figure~\ref{fig:selection} reports category-level selection accuracy for Llama-3.2-1B and Qwen2.5-3B. Both of our methods outperform ARROW on every backbone and every category. ARIADNE reaches an average selection accuracy of 0.85, and remains stable as the library grows, reaching 0.897 across 44 adapters (Scalability to 44 Tasks Section of Appendix). \method{} reaches 0.4174 on Llama-3.2-1B and 0.5548 on Qwen2.5-3B, against ARROW's 0.0625 and 0.0834 respectively. The advantage of \method{} over ARROW is statistically robust on every backbone (paired t-test, all $p<0.005$; Wilcoxon signed-rank $p<0.0001$ on Qwen2.5-3B), and ARROW's selection accuracy remains close to its random-choice baseline throughout (0.0435), failing to significantly exceed it on any individual backbone (one-sample Wilcoxon test, $p=0.052$--$0.100$). This quantitatively corroborates previous observations that ARROW's per-token, per-layer selection is close to random~\cite{fleshman2025spectr}. The gap between our two methods is the price of the training data. Removing it costs roughly half of ARIADNE's selection accuracy on Llama-3.2-1B and a third on Qwen2.5-3B, and the loss is not uniform across categories: \method{} retains most of ARIADNE's accuracy on Similarity (0.69 and 0.93 vs.\ 0.96) and Reasoning (0.67 vs.\ 1.00), but degrades sharply on NLI (0.22 and 0.38 vs.\ 0.81). This is the expected failure mode of extraction-based characterization. As reported in the Qualitative Results Section, the \textsc{rte}, \textsc{anli\_r1}, and \textsc{snli} adapters all reconstruct the \emph{same} training template verbatim, so their synthetic pools are near-identical and their centroids collapse onto one another. Where the NLI family is concerned, CDD-UM recovers the task format faithfully but cannot distinguish adapters that share it.

\subsubsection{Task Accuracy.}

\begin{figure}[ht]
    \centering
    \includegraphics[width=\columnwidth]{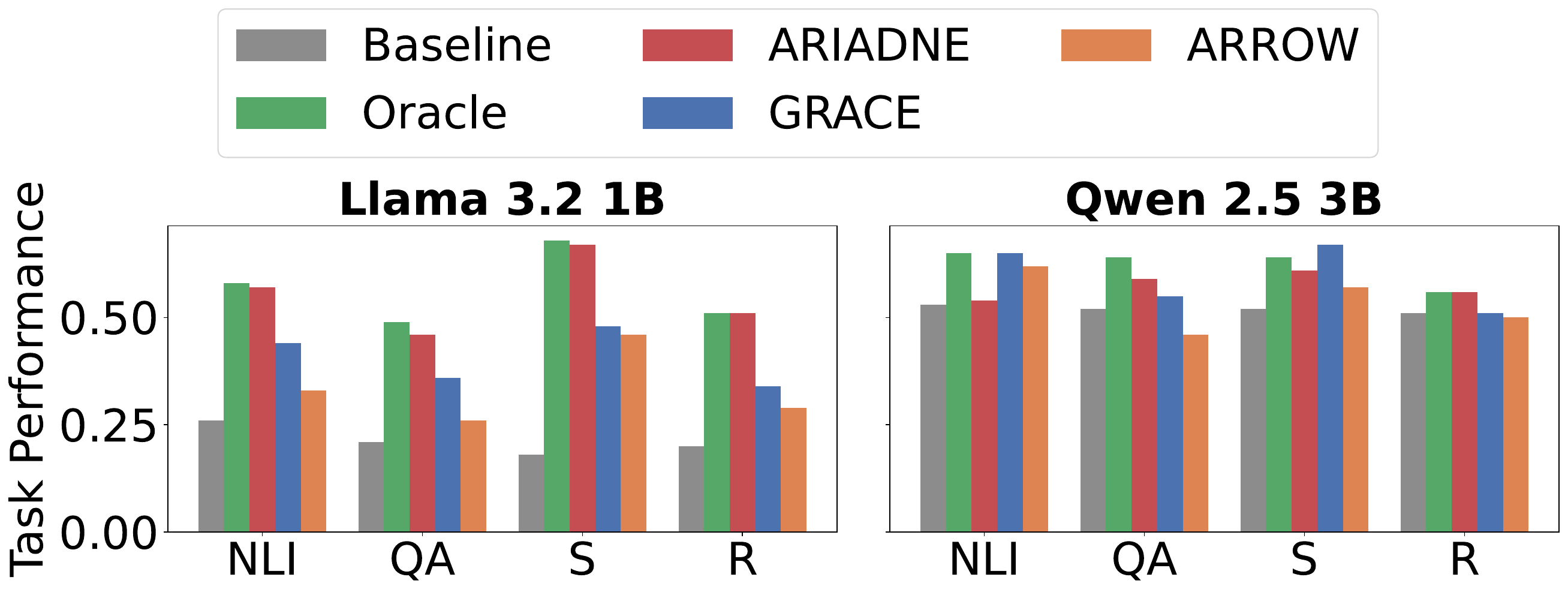}
    \caption{Category-level task accuracy (Baseline vs.\ Oracle vs.\ ARIADNE vs.\
    \method{} vs.\ ARROW) on Llama-3.2-1B and Qwen2.5-3B for Natural Language Inference (NLI), Question Answering (QA), Similarity (S), and Reasoning (R). ARIADNE closes almost the
    entire gap to Oracle on both backbones. \method{} reaches 72\% of Oracle on
    Llama-3.2-1B and 95.6\% on Qwen2.5-3B, where it slightly exceeds ARIADNE despite
    using no training data. Both methods outperform ARROW on average.}
\label{fig:tasks}
\end{figure}

Figure~\ref{fig:tasks} reports category-level task accuracy. ARIADNE recovers 97.4\% of Oracle on Llama-3.2-1B (0.55 vs.\ 0.57) and 90.2\% on Qwen2.5-3B (0.58 vs.\ 0.63). \method{} recovers 72\% of Oracle on Llama-3.2-1B (0.41 vs.\ 0.57) and 95.6\% on Qwen2.5-3B (0.60 vs.\ 0.63). Against ARROW, \method{} matches or exceeds accuracy on 16/23 and 17/23 tasks respectively, and its average exceeds ARROW's on both backbones; a paired Wilcoxon signed-rank test confirms the advantage is significant ($p = 0.003$ pooled across all task-backbone combinations). For the full tables, see Appendix Section Full Per-Task Results. The comparison between our two methods is where the interesting result lies. The selection-accuracy gap of Figure~\ref{fig:selection} does not transfer proportionally to task accuracy. On Llama-3.2-1B, \method{} trails ARIADNE by 0.14 in task accuracy despite trailing by more than twice that in selection accuracy. On Qwen2.5-3B the ordering reverses outright: \method{} attains 0.60 against ARIADNE's 0.58, outperforming on task accuracy the very method whose selection accuracy is nearly twice its own. This is consistent with the graceful degradation reported for ARIADNE (Analysis of Routing Failures section in Appendix): because routing happens in a semantically structured embedding space, misroutes tend to land on adapters trained for closely related objectives, which absorb much of the error.


\subsubsection{Scale Extension: Qwen2.5-32B.}
\label{sec:qwen32b}

\begin{figure}[ht]
    \centering
    \includegraphics[width=\columnwidth]{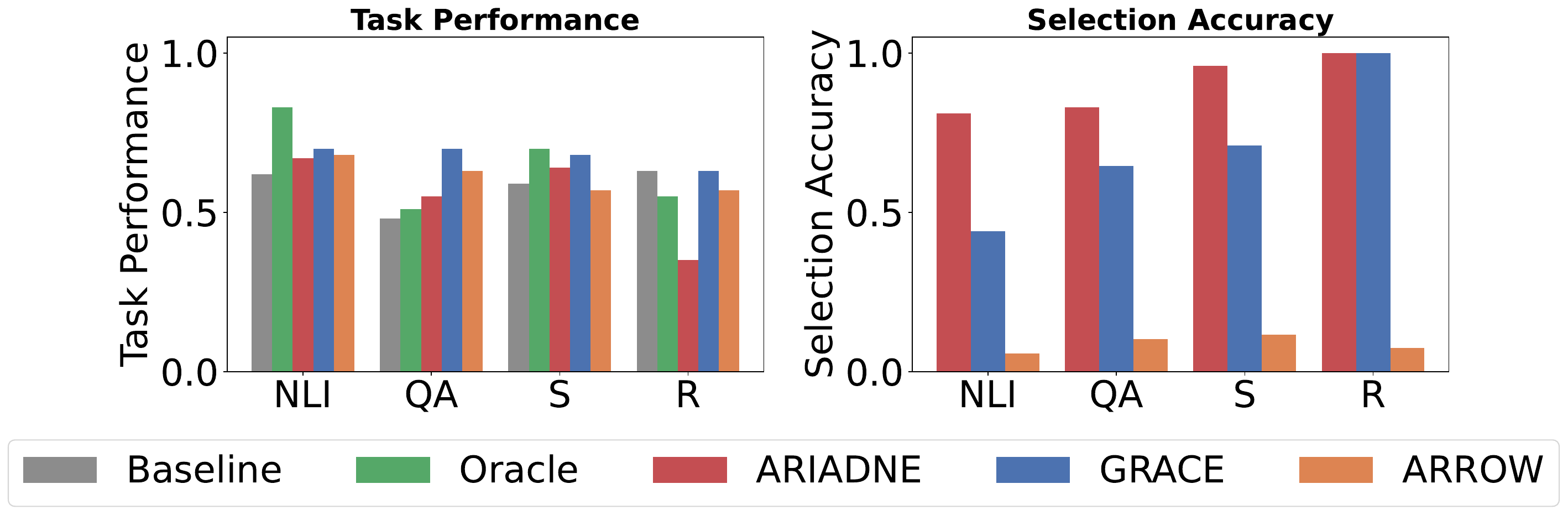}
    \caption{Task accuracy (left) and selection accuracy (right) on Qwen2.5-32B-Instruct for Natural Language Inference (NLI), Question Answering (QA), Similarity (S), and Reasoning (R).
    \method{} attains its highest selection accuracy of any backbone (0.6226 vs.\ ARROW's
    0.0835) and exceeds Oracle in task accuracy (0.68 vs.\ 0.65), driven by QA and
    Reasoning, where the task-matched adapter underperforms both \method{}'s choice and,
    on Reasoning, the unadapted Baseline.}
    \label{fig:qwen32b}
\end{figure}

To verify that the conclusions hold at scale, we repeat the evaluation on Qwen2.5-32B-Instruct, an order of magnitude larger than our second backbone (Figure~\ref{fig:qwen32b}).  \method{} improves with scale on both metrics, reaching a selection accuracy of 0.6226 against ARROW's 0.0835, the largest margin of any backbone, with high selection accuracy on the Reasoning category. In task accuracy it attains 0.68, exceeding Oracle's 0.65. The category breakdown makes clear why this is possible rather than paradoxical: on QA, \method{} reaches 0.70 against Oracle's 0.51, and on Reasoning, Oracle (0.55) falls below the unadapted Baseline (0.63) altogether.

\subsubsection{First token validation of CDD-UM}

We further verify the structural motivation behind CDD-UM by probing the next-token distribution immediately after the user-turn opening tag (Fine-Tuning Signal at the First User Token Section of Appendix). Fine-tuning typically promotes the first token of the task prompt from a relatively unlikely base-model continuation to the top-ranked prediction, often with near-unit probability. This confirms that CDD-UM begins generation at a position where the prompt-template signal is already directly exposed, rather than relying only on autoregressive self-reinforcement.
\section{Discussion}
\label{sec:discussion}

\textbf{Weight-free by construction.} Neither of our methods inspects an adapter parameter, activation, or gradient at any point. ARIADNE sees only training inputs; \method{} sees only the final-layer logits of $p_{\mathrm{base}}$ and $p_{\mathrm{ft}}^{(i)}$ during its offline pass. This design choice has two consequences we discuss in turn: it makes the pipeline agnostic to the underlying PEFT architecture, and it reflects a deliberate skepticism toward relying on model internals at all.

\textbf{Architecture independence.} ARROW constructs routing prototypes from the singular value decomposition of the LoRA weight update $\Delta W = BA$. This construction presupposes that $\Delta W$ is available as an explicit low-rank factorization, which holds for LoRA but not for other widely used PEFT families: BitFit~\citep{zaken2022bitfit} updates only bias terms and produces no weight-update matrix; IA3~\citep{liu2022ia3} rescales activations with learned vectors rather than producing an additive $\Delta W$; and full finetuning produces a dense, typically full-rank $\Delta W$, for which an SVD carries none of the low-rank structure ARROW's prototype depends on. Adapting ARROW to any of these settings would require redesigning the prototype-construction step around each architecture's own parameterization, and for some (IA3, full finetuning) it is not obvious such a redesign exists at all. Neither of our methods requires such redesign. ARIADNE never consults the adapter at all: its centroids are built from task inputs, so the adapter could be of any form whatsoever, and the same holds for any future PEFT method not yet invented. \method{} characterizes an adapter purely through the input-output behavior, so the underlying PEFT method is likewise invisible to the pipeline: the same offline procedure applies unchanged whether $\phi_i$ is a LoRA adapter, a BitFit bias vector, an IA3 rescaling, or a fully finetuned checkpoint. The requirement is not on the adapter's form but only on the ability to query the adapted model's output distribution.

\textbf{The $GL(r)$ non-uniqueness of LoRA is not what distinguishes us from ARROW.} The bilinear factorization $\Delta W = BA$ is non-unique: for any invertible $G \in GL(r)$, the substitution $B \mapsto BG^{-1}$, $A \mapsto GA$ leaves $\Delta W$ unchanged, so any quantity computed directly from the raw $(B,A)$ pair is a property of the representation, not of the underlying adapter~\citep{mandica2026gpartendtoendisometricfinetuning}. This is a real issue for methods that route or merge adapters using the raw factors: LEGO~\citep{zhao2025merging} clusters LoRA ranks by Euclidean distance between the concatenated vectors $[a_i, b_i]$, and HiLoRA's token-level routing~\citep{han2025hilora} ranks rank-one components by the projection $a_{ij}^\top x$ computed from the $A$-row alone; both scores change under the rescaling $a_i \mapsto \lambda a_i$, $b_i \mapsto \lambda^{-1} b_i$ that leaves the adapter's contribution to $\Delta W$ exactly unchanged. ARROW avoids this: it canonicalizes through the SVD of $\Delta W = U\Sigma V^\top$ before constructing prototypes, which \citet{mandica2026gpartendtoendisometricfinetuning} characterize as a principled and largely complete fix for this non-uniqueness when singular values are distinct, the generic case for trained adapters. The actual distinguishing factor in our comparison is therefore not this pathology, but the access requirement discussed above: ARROW still needs white-box access to $\Delta W$ at prototype-construction time and is tied to architectures that produce an explicit weight-update matrix to decompose, whereas neither of our methods requires either.

\textbf{Avoiding model internals is deliberate, not merely convenient.} Our motivation for routing in embedding space rather than weight space is partly that representation-based signals offer no guarantee of correspondence with the structure they are probed for~\citep{fleshman2025spectr, dubanowska-etal-2025-representation}; \citet{dubanowska-etal-2025-representation} show specifically that representation-based hallucination detectors, trained on a model's internal activations, fail to generalize out of distribution. \method{} extends this skepticism from activation probes to mechanistic-interpretability tooling more broadly. CDD itself was developed in contrast to exactly such a white-box alternative: \citet{brzozowski2026readingfinetuningpriorverbatim} report that grey-box CDD, reading only output logits, recovers verbatim finetuning content more reliably than a white-box activation-difference method (ADL) with full internal access to the same models. Restricting \method{} to logits is not only what makes it architecture-agnostic, it also sidesteps a class of internal signals whose reliability is unsettled.

\section{Conclusion} We presented ARIADNE, a training-free adapter routing framework that represents each adapter by centroids in the latent space of a frozen text encoder, and \method{}, a grey-box instantiation that needs neither the training data nor the model weights. Unlike the state of the art ARROW \citep{ostapenko2024towards}, both avoid weight-space operations completely. By synthesizing adapter-specific training prompts from grey-box logit access alone with our CDD-UM, \method{} additionally eliminates ARIADNE's training-data requirement. In both methods the online router reduces to a nearest-centroid lookup and requires no model access after the offline characterization phase. Beyond the empirical results, this work demonstrates that model diffing, a tool developed for mechanistic interpretability, can serve as a practical primitive in deployment pipelines. Staying out of weight space throughout, where the geometry is shaped by a parameterization rather than by the task, yields routers that require no access to adapter weights and are therefore agnostic to the underlying PEFT architecture, a property weight-space methods cannot offer by construction. 

\textbf{Limitations.} CDD-UM cannot separate adapters that share a training template, which is why \method{}'s selection accuracy is lowest on the NLI family, and sentence quality may further degrade for very low-rank or heavily quantized adapters; we leave a systematic study of this regime to future work. The offline phase costs one CDD pass per adapter, which is amortized at deployment but non-negligible during library construction. 

\textbf{Future work.} Natural extensions include applying the pipeline to non-LoRA PEFT methods such as IA3 and prefix tuning, combining our routers with GPart adapters that eliminate $BA$ pathologies by construction, and exploring whether CDD-derived sentence distributions can guide adapter merging and interpolation. Resolving the template-collapse failure mode is the most direct route to closing the remaining gap between the two methods.


\bibliography{aaai2027}

\newpage
\appendix
\clearpage
\section{Appendix}

\subsection{Full Per-Task Results}
\label{sec:full_results_em}

Tables~\ref{tab:full_em} and~\ref{tab:full_em_qwen} report, for each task, the performance of the unadapted backbone (Base), the Oracle (the base model with the adapter trained on that specific task), and the base model with the adapter selected dynamically by ARIADNE and by \method{}. For all tasks the metric is Exact Match, with the exception of CommonGen, which is measured with ROUGE. These tables expand the category-level results of Figure~\ref{fig:tasks}.
\begin{table}[ht]
    \centering
    \small
    \begin{tabular}{lcccc}
    \toprule
    Dataset & Base & Oracle & ARIADNE & \method{} \\
    \midrule
    WSC & 0.02 & 0.51 & 0.51 & 0.46 \\
    DPR & 0.34 & 0.62 & 0.62 & 0.45 \\
    PAWS WIKI & 0.46 & 0.90 & 0.88 & 0.78 \\
    Trivia QA & 0.14 & 0.30 & 0.30 & 0.14 \\
    QQP & 0.00 & 1.00 & 1.00 & 0.31 \\
    MRPC & 0.26 & 0.70 & 0.70 & 0.69 \\
    STS-B & 0.00 & 0.11 & 0.11 & 0.12 \\
    CB & 0.32 & 0.43 & 0.43 & 0.32 \\
    WNLI & 0.42 & 0.44 & 0.41 & 0.48 \\
    ANLI R1 & 0.16 & 0.38 & 0.42 & 0.28 \\
    ANLI R2 & 0.21 & 0.44 & 0.39 & 0.41 \\
    ANLI R3 & 0.24 & 0.44 & 0.43 & 0.36 \\
    MNLI Matched & 0.19 & 0.85 & 0.72 & 0.37 \\
    MNLI Mismatched & 0.20 & 0.86 & 0.86 & 0.32 \\
    SNLI & 0.30 & 0.88 & 0.88 & 0.91 \\
    RTE & 0.28 & 0.56 & 0.56 & 0.54 \\
    CommonGen & 0.25 & 0.41 & 0.41 & 0.11 \\
    SQuAD V1 & 0.54 & 0.75 & 0.64 & 0.55 \\
    SQuAD V2 & 0.17 & 0.65 & 0.62 & 0.25 \\
    BoolQ & 0.37 & 0.71 & 0.70 & 0.70 \\
    NQ & 0.03 & 0.11 & 0.11 & 0.14 \\
    ARC-Easy & 0.11 & 0.56 & 0.54 & 0.45 \\
    ARC-Challenge & 0.11 & 0.33 & 0.33 & 0.31 \\
    \midrule
    Average & 0.22 & 0.56 & 0.55 & 0.41 \\
    \bottomrule
    \end{tabular}
    \caption{Per-task performance on Llama-3.2-1B-Instruct. Metric is Exact Match
    except for CommonGen (ROUGE).}
    \label{tab:full_em}
\end{table}

\begin{table}[ht]
    \centering
    \small
    \begin{tabular}{lcccc}
    \toprule
    Dataset & Base & Oracle & ARIADNE & \method{} \\
    \midrule
    WSC & 0.60 & 0.62 & 0.62 & 0.61 \\
    DPR & 0.62 & 0.62 & 0.62 & 0.58 \\
    PAWS WIKI & 0.74 & 0.90 & 0.83 & 0.82 \\
    Trivia QA & 0.16 & 0.20 & 0.20 & 0.24 \\
    QQP & 0.70 & 1.00 & 1.00 & 1.00 \\
    MRPC & 0.64 & 0.64 & 0.60 & 0.73 \\
    STS-B & 0.00 & 0.02 & 0.02 & 0.11 \\
    CB & 0.34 & 0.28 & 0.24 & 0.43 \\
    WNLI & 0.34 & 0.34 & 0.32 & 0.42 \\
    ANLI R1 & 0.54 & 0.66 & 0.30 & 0.60 \\
    ANLI R2 & 0.34 & 0.44 & 0.17 & 0.48 \\
    ANLI R3 & 0.26 & 0.58 & 0.55 & 0.43 \\
    MNLI Matched & 0.70 & 0.88 & 0.63 & 0.88 \\
    MNLI Mismatched & 0.76 & 0.90 & 0.90 & 0.89 \\
    SNLI & 0.88 & 0.92 & 0.92 & 0.94 \\
    RTE & 0.58 & 0.86 & 0.86 & 0.75 \\
    CommonGen & 0.31 & 0.44 & 0.44 & 0.35 \\
    SQuAD V1 & 0.72 & 0.80 & 0.73 & 0.72 \\
    SQuAD V2 & 0.24 & 0.66 & 0.59 & 0.38 \\
    BoolQ & 0.70 & 0.86 & 0.72 & 0.82 \\
    NQ & 0.10 & 0.16 & 0.16 & 0.11 \\
    ARC-Easy & 0.86 & 0.90 & 0.83 & 0.79 \\
    ARC-Challenge & 0.84 & 0.90 & 0.90 & 0.79 \\
    \midrule
    Average & 0.52 & 0.63 & 0.57 & 0.60 \\
    \bottomrule
    \end{tabular}
    \caption{Per-task performance on Qwen2.5-3B-Instruct. Metric is Exact Match
    except for CommonGen (ROUGE).}
    \label{tab:full_em_qwen}
\end{table}

\begin{table}[ht]
    \centering
    \small
    \begin{tabular}{lcccc}
    \toprule
    Dataset & Base & Oracle & ARIADNE & \method{} \\
    \midrule
    WSC & 0.74 & 0.62 & 0.62 & 0.62 \\
    DPR & 0.90 & 0.92 & 0.92 & 0.90 \\
    PAWS WIKI & 0.82 & 0.94 & 0.88 & 0.86 \\
    Trivia QA & 0.16 & 0.16 & 0.16 & 0.30 \\
    QQP & 0.70 & 1.00 & 1.00 & 1.00 \\
    MRPC & 0.80 & 0.86 & 0.83 & 0.84 \\
    STS-B & 0.02 & 0.00 & 0.00 & 0.00 \\
    CB & 0.36 & 0.82 & 0.30 & 0.34 \\
    WNLI & 0.12 & 0.88 & 0.32 & 0.80 \\
    ANLI R1 & 0.80 & 0.84 & 0.78 & 0.86 \\
    ANLI R2 & 0.56 & 0.66 & 0.54 & 0.62 \\
    ANLI R3 & 0.54 & 0.74 & 0.58 & 0.58 \\
    MNLI Matched & 0.82 & 0.88 & 0.81 & 0.76 \\
    MNLI Mismatched & 0.82 & 0.86 & 0.86 & 0.58 \\
    SNLI & 0.82 & 0.88 & 0.88 & 0.88 \\
    RTE & 0.78 & 0.94 & 0.94 & 0.90 \\
    CommonGen & 0.26 & 0.10 & 0.10 & 0.37 \\
    SQuAD V1 & 0.24 & 0.28 & 0.22 & 0.86 \\
    SQuAD V2 & 0.10 & 0.12 & 0.61 & 0.60 \\
    BoolQ & 0.82 & 0.90 & 0.86 & 0.88 \\
    NQ & 0.10 & 0.12 & 0.12 & 0.26 \\
    ARC-Easy & 0.96 & 0.98 & 0.90 & 0.98 \\
    ARC-Challenge & 1.00 & 1.00 & 1.00 & 1.00 \\
    \midrule
    Average & 0.58 & 0.67 & 0.62 & 0.69 \\
    \bottomrule
    \end{tabular}
    \caption{Per-task performance on Qwen2.5-32B-Instruct. Metric is Exact Match except for CommonGen (ROUGE).}
    \label{tab:full_em_qwen32b_merged}
\end{table}

%
%

\subsection{Fine-Tuning Signal at the First User Token}
\label{sec:first-token-probe}

To verify that the user-turn opening tag exposes the prompt-template signal
targeted by CDD-UM, we probe the next-token distribution immediately after
this tag. For each Qwen2.5-3B adapter, we take the first token of its
fine-tuning task prompt as the target token and measure its probability and
rank under the base model, the fine-tuned model, and contrastive decoding.
This isolates the first decoding step, before any generated tokens can
reinforce the task template autoregressively.

Across the 23 available adapters, the target token is never the top-ranked
continuation of the base model, but becomes rank~1 under the fine-tuned model
for 20 adapters. The median target-token probability increases from
$0.0365$ under the base model to $0.9999$ after fine-tuning. The effect is
particularly pronounced when the task template begins with a distinctive
token. For example, the probability of \texttt{Answer} rises from $0.0011$
to above $0.9999$ for both TriviaQA and Natural Questions, while
\texttt{Read} rises from $0.0015$ to $0.9995$ for BoolQ. Representative
results are shown in Table~\ref{tab:first-token-probe}.

\begin{table}[t]
    \centering
    \small
    \begin{tabular}{lcrrr}
        \hline
        Task & Token & $p_{\mathrm{base}}$ & $p_{\mathrm{ft}}$
             & $p_{\mathrm{CDD}}$ \\
        \hline
        TriviaQA      & \texttt{Answer}   & 0.0011 & 1.0000 & 1.0000 \\
        BoolQ         & \texttt{Read}     & 0.0015 & 0.9995 & 0.9957 \\
        ARC-Challenge & \texttt{Given}    & 0.0365 & 0.1028 & 0.1913 \\
        PAWS-Wiki     & \texttt{Given}    & 0.0365 & 0.9999 & 1.0000 \\
        \hline
    \end{tabular}
    \caption{Probability assigned to the first token of the fine-tuning task
    prompt immediately after the user-turn opening tag for Qwen2.5-3B.
    $p_{\mathrm{CDD}}$ uses contrastive decoding with $\beta=1$ and no
    plausibility filtering. Values are rounded to four decimal places.}
    \label{tab:first-token-probe}
\end{table}

These results provide direct evidence for the structural choice made by
CDD-UM: fine-tuning deposits a strong prompt-template signal precisely at the
beginning of the user turn. Contrastive decoding can further amplify this
signal, although unconstrained amplification is not monotonic in $\beta$.
For some adapters, tokens with negligible base-model probability attain a
larger likelihood ratio and overtake the true first token at high $\beta$.
The plausibility constraint used by CDD-UM suppresses this failure mode.
Overall, the probe shows that prompt recovery begins with an explicit
fine-tuning signal at the first decoding step rather than arising only from
autoregressive continuation.

\subsection{Embedder Search}
\label{sec:embedder_selection}

Due to the large number of existing text embedders, our selection is based upon a search
on a subset of tasks. We pick as most suitable encoder the one that yields the highest
cosine similarity between the generated task representation, i.e.\ the centroids, and 20
test set samples. We perform this study on the GLUE tasks~\cite{wang2018glue}; results are
reported in Table~\ref{tab:appendix_embedders}. The selected encoder,
\texttt{intfloat/e5-large-v2}, is used unchanged by both methods, so that the comparison
between them isolates the provenance of the characterization sentences rather than the
embedding space they are projected into.

\begin{table}[t]
    \centering
    \small
    \begin{tabular}{lccccc}
        \toprule
        \textbf{Model} & \textbf{CoLA} & \textbf{RTE} & \textbf{MRPC} & \textbf{SST-2} & \textbf{Avg.} \\
        \midrule
        MiniLM$^\dagger$      & 0.111 & 0.090 & 0.271 & 0.433 & 0.226 \\
        Qwen3-0.6B$^\ddagger$ & 0.532 & 0.400 & 0.415 & 0.686 & 0.508 \\
        mE5-small$^\S$        & 0.860 & 0.848 & 0.868 & 0.881 & 0.864 \\
        E5-large-v2$^\P$      & \textbf{0.910} & \textbf{0.877} & \textbf{0.902} & \textbf{0.891} & \textbf{0.895} \\
        \bottomrule
    \end{tabular}
    \caption{Similarity scores across GLUE tasks centroids and test
    samples. Bold = best, justifying our embedder choice.
    $^\dagger$\texttt{all-MiniLM-L12-v2};
    $^\ddagger$\texttt{Qwen/Qwen3-Embedding-0.6B}~\cite{qwen3embedding};
    $^\S$\texttt{intfloat/multilingual-e5-small}~\cite{wang2024multilingual};
    $^\P$\texttt{intfloat/e5-large-v2}~\cite{wang2024multilingual}.}
    \label{tab:appendix_embedders}
\end{table}

\subsection{Number of Centroids}
\label{sec:centroids_number}

\begin{figure*}
    \centering
    \includegraphics[width=.8\linewidth]{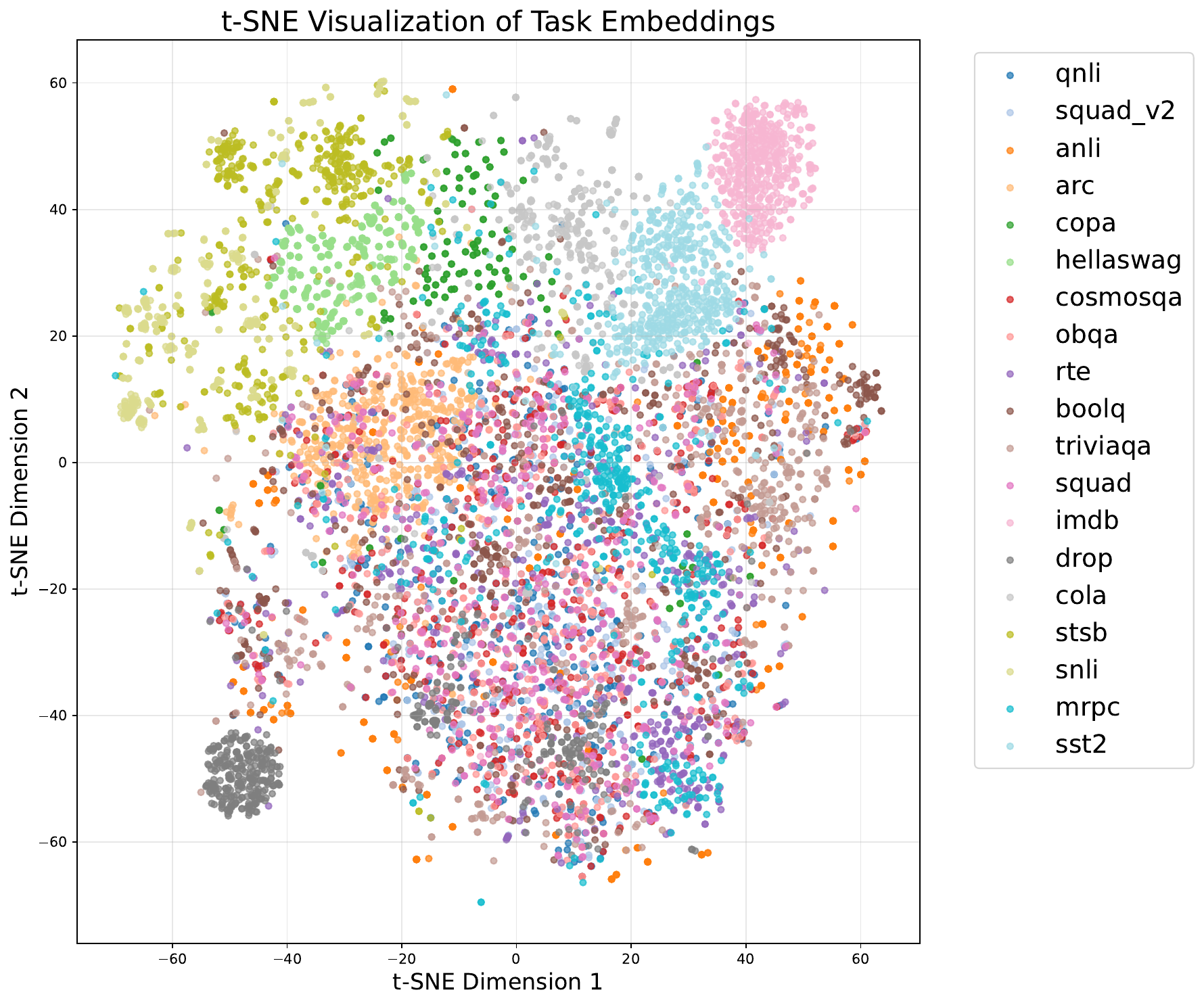}
    \caption{t-SNE analysis of the task embeddings. }
    \label{fig:tsne}
\end{figure*}

To choose the number of centroids we measured selection accuracy on a subset of 19 tasks
under three configurations (Table~\ref{tab:ablation_centroids}). The multi-centroid
formulation is motivated by the intra-task embedding variance and illustrated in Figure~\ref{fig:tsne}: adversarially
constructed tasks such as ANLI produce inputs whose embeddings span multiple disjoint
regions of the latent space, while tasks with heterogeneous input formats, such as
extractive vs.\ abstractive question answering, may cluster around semantically distinct
centroids even within the same task. A single global mean collapses this structure and
produces a centroid that may lie in a low-density region, making it a poor representative
of any individual input. K-NN routing, by contrast, lacks task-level structure entirely
and conflates inter-task proximity with intra-task variance, explaining its intermediate
performance. We adopt $m=5$ for both methods.

\begin{table}[ht]
\centering
\small
\begin{tabular}{l|ccc}
\toprule
\textbf{Metric} & \textbf{1 Centroid} & \textbf{5 Centroids} & \textbf{K-NN} \\
\midrule
SA & 47.4\% & 89.4\% & 65.9\% \\
\bottomrule
\end{tabular}
\caption{Selection accuracy over 19 tasks under three routing configurations. The
sweetspot lies between a single centroid per task and no centroid structure at all
(K-NN).}
\label{tab:ablation_centroids}
\end{table}

\subsection{Analysis of Routing Failures}
\label{sec:failures}

We identify three failure modes shared by both methods, and one specific to \method{}.

\textit{(1) Domain overlap.} SQuAD V1 achieves 0\% selection accuracy under ARIADNE, being
consistently misrouted to the SQuAD V2 adapter. The system nevertheless recovers 85\% of
Oracle performance (0.64 vs.\ 0.75 Task Accuracy), demonstrating that semantically proximate adapters
absorb routing errors.

\textit{(2) Adversarial variance.} ANLI R1 and R2 exhibit lower selection accuracy (0.46
and 0.38) due to high intra-task embedding variance arising from their adversarial
construction. ANLI R1's downstream task performance actually exceeds the Oracle's (0.42
vs.\ 0.38), suggesting beneficial cross-task generalization from other NLI adapters.

\textit{(3) Reasoning ambiguity.} Complex tasks such as MultiRC are frequently misrouted to
general QA centroids (TriviaQA) due to shared linguistic surface features.

\textit{(4) Template collapse (\method{} only).} Adapters trained on the same prompt
template produce near-identical synthetic pools, and hence near-identical centroids. The
NLI family is the clearest instance: \textsc{rte}, \textsc{anli\_r1}, and \textsc{snli} all
reconstruct the same instruction verbatim, which is why
\method{}'s selection accuracy is lowest in that category. Unlike modes (1)--(3), this
failure originates in the offline phase rather than in the geometry of the embedding space.

\begin{figure*}
    \centering
    \includegraphics[width=.8\linewidth]{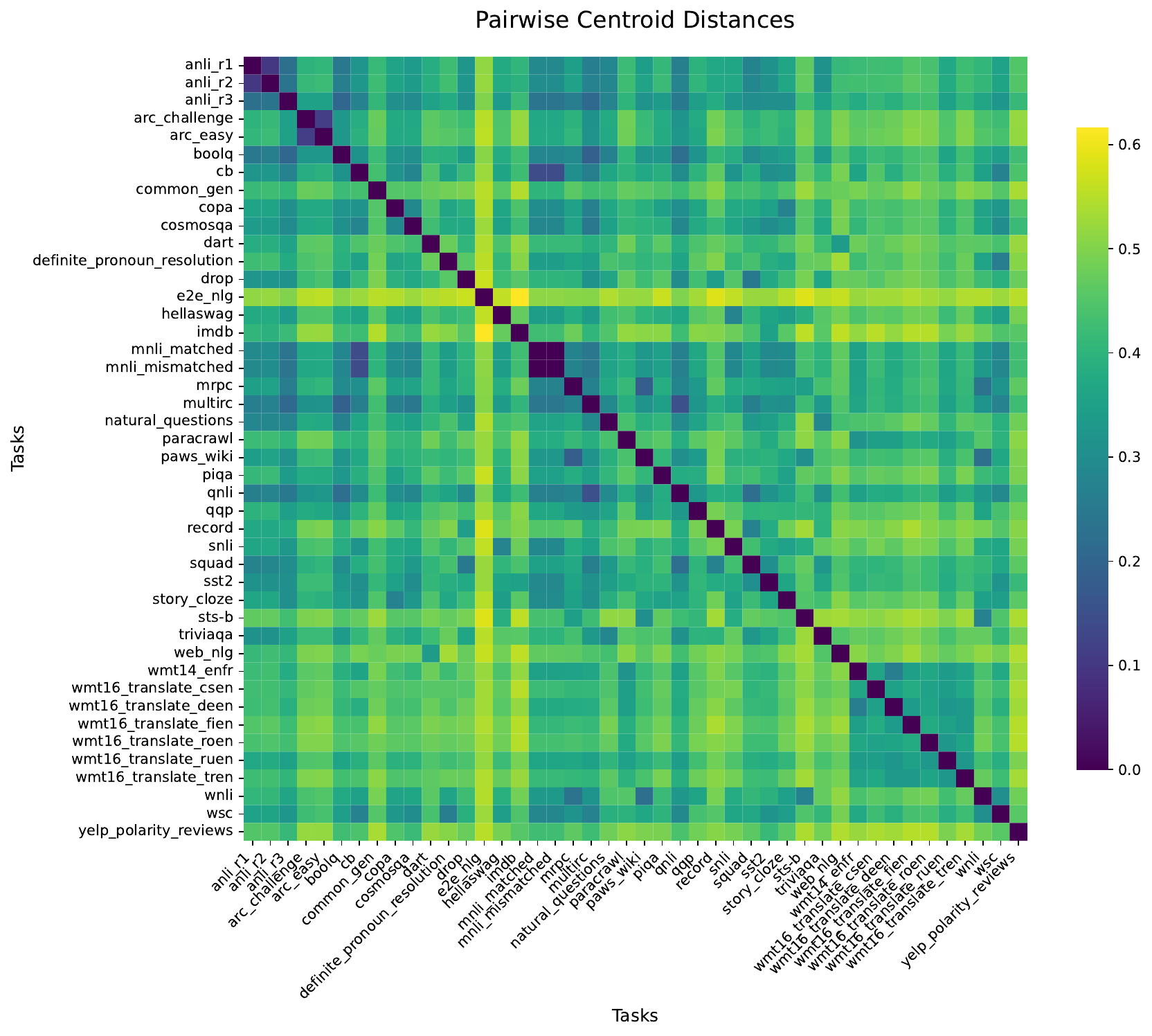}
    \caption{Pairwise distances between task centroids. Visual interpretation of the graceful degradation property.}
    \label{fig:distances}
\end{figure*}

Across all modes, errors concentrate within the same semantic cluster, producing graceful
rather than catastrophic degradation. Figure~\ref{fig:distances} makes this visible:
semantically similar tasks achieve the highest centroid similarity, so a misroute lands on
an adapter trained for a closely related objective.

\subsection{Characterization Set Size}
\label{sec:centroid_num_samples}

\begin{figure}
    \centering
    \includegraphics[width=\linewidth]{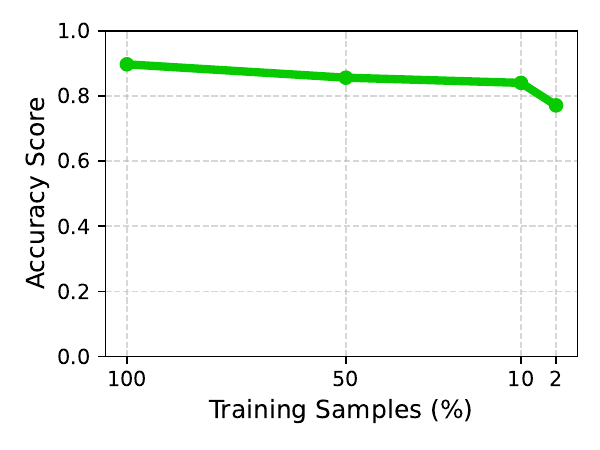}
    \caption{Selection accuracy as a function of the number of characterization sentences per adapter. ARIADNE's best performance is achieved with 500 training samples, as reported in the main paper, and remains strong down to 2\%.}
    \label{fig:centroid_num_samples}
\end{figure}

Both methods summarize a pool of sentences into centroids, and the size of that pool is a free parameter. Reducing the number of training samples available to ARIADNE has only a mild effect on selection accuracy (Figure~\ref{fig:centroid_num_samples}): with 2\% of the original samples, SA is still 77.1\%. We perform this study only for ARIADNE, since it is the method for which pool size is a binding constraint: an adapter's training set may be small, partially released, or otherwise restricted, so it matters how far selection accuracy holds up as the available sample count shrinks. The question does not arise for \method{}, whose pool is generated rather than given, and can therefore be enlarged to an arbitrary size at the cost of additional CDD-UM generations alone. We fix $|\hat{S}_i| = 3000$ throughout.

\subsection{Scalability to 44 Tasks}
\label{sec:full_routing}

\begin{figure}
    \centering
    \includegraphics[width=.8\linewidth]{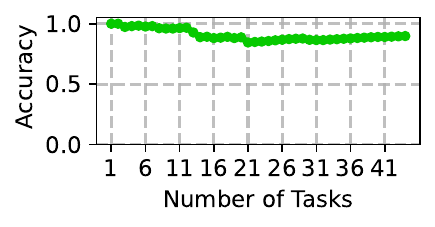}
    \caption{Selection Accuracy trend for up to 44 tasks.}
    \label{fig:scalability_trend}
\end{figure}

To evaluate whether routing precision degrades as the adapter library grows, we extend the
evaluation to a full set of 44 tasks, adding 21 to the primary evaluation.
The accuracy scalability trend is reported in Figure~\ref{fig:scalability_trend}.

Table~\ref{tab:scalability} reports per-task selection accuracy. Selection accuracy
stabilizes at 89.7\% and plateaus beyond roughly 20 adapters rather than continuing to
degrade, consistent with the routing failure analysis: errors
concentrate within semantic clusters, so enlarging the pool with distant tasks adds few new
confusions. This study is run with ARIADNE, since it requires only the training sets of the
21 additional tasks and no additional extraction passes.
\begin{table}[ht]
\centering
\small
\begin{tabular}{lc}
\toprule
Dataset & Acc. \\
\midrule
\multicolumn{2}{l}{\textit{Primary evaluation (23 tasks)}} \\
\midrule
WSC                 & 1.00 \\
DPR                 & 1.00 \\
PAWS Wiki           & 0.92 \\
Trivia QA           & 1.00 \\
QQP                 & 1.00 \\
MRPC                & 0.94 \\
STS-B               & 1.00 \\
CB                  & 0.84 \\
WNLI                & 0.94 \\
BoolQ               & 0.96 \\
NQ                  & 1.00 \\
ARC-Challenge       & 1.00 \\
ANLI R1             & 0.46 \\
ANLI R2             & 0.38 \\
ANLI R3             & 0.94 \\
MNLI Matched        & 0.72 \\
MNLI Mismatched     & 1.00 \\
SNLI                & 1.00 \\
RTE                 & 1.00 \\
CommonGen           & 1.00 \\
SQuAD V1            & 0.00 \\
SQuAD V2            & 0.94 \\
ARC-Easy            & 0.92 \\
\midrule
\multicolumn{2}{l}{\textit{Scalability extension (21 tasks)}} \\
\midrule
WMT16 Ro-En         & 0.96 \\
DART                & 1.00 \\
DROP                & 0.70 \\
ParaCrawl           & 0.80 \\
Story Cloze         & 1.00 \\
HellaSwag           & 1.00 \\
PIQA                & 0.98 \\
WMT16 Fi-En         & 1.00 \\
WMT16 Tr-En         & 1.00 \\
WMT16 Ru-En         & 1.00 \\
WMT16 De-En         & 1.00 \\
IMDB                & 0.96 \\
WMT14 En-Fr         & 0.92 \\
MultiRC             & 0.56 \\
E2E NLG             & 1.00 \\
ReCoRD              & 0.94 \\
SST-2               & 0.98 \\
WMT16 Cs-En         & 1.00 \\
Yelp Reviews        & 0.86 \\
COPA                & 1.00 \\
QNLI                & 0.88 \\
\midrule
Average (44 tasks)  & 0.897 \\
\bottomrule
\end{tabular}
\caption{ARIADNE adapter selection accuracy across all 44 tasks. The upper block
corresponds to the 23 primary evaluation tasks; the lower block constitutes the
scalability extension. The average is computed over all 44 tasks.}
\label{tab:scalability}
\end{table}

\subsection{Inference Cost}
\label{sec:inference_costs}
Since both methods choose an adapter at inference time, it is important to quantify the
associated overhead. The embedding step, which uses \texttt{intfloat/e5-large-v2}, takes on
average 20.04 ms with a standard deviation of $\pm3.70$ ms. The subsequent selection of the
most appropriate adapter, based on the pre-computed centroids, adds another 1.98 ms
($\pm0.10$ ms). The complete selection pipeline therefore incurs a total latency of roughly
22.02 ms ($\pm3.66$ ms), after which inference proceeds as usual by applying the chosen
adapter to the base model. Because the online phase is identical in the two methods, these figures apply to both: whatever was done offline to
obtain the centroids, the router itself is the same nearest-centroid lookup over the same
number of centroids, and contributes nothing beyond the latency reported above.

\end{document}